\documentclass[letterpaper]{article} 
\usepackage{aaai2026}  
\usepackage{times}  
\usepackage{helvet}  
\usepackage{courier}  
\usepackage[hyphens]{url}  
\usepackage{graphicx} 
\usepackage{natbib}  
\usepackage{caption} 
\usepackage{algorithm}
\usepackage{algorithmic}
\usepackage{microtype}
\usepackage{graphicx}
\usepackage{subfigure}
\usepackage{multirow}
\usepackage{booktabs} 
\usepackage{amsmath} 
\usepackage{pgfplots}
\pgfplotsset{compat=1.18}
\usepackage{makecell}
\usepackage{enumitem}
\usepackage{booktabs}
\usepackage{graphicx}
\usepackage{array}
\usepackage{soul}
\usepackage{url}
\usepackage{xcolor}
\usepackage{booktabs}
\usepackage[table]{xcolor}
\usepackage{array}
\usepackage{caption}
\usepackage{booktabs}
\usepackage{tabularx}
\usepackage{tikz}
\usetikzlibrary{positioning,arrows.meta}
\usepackage{booktabs}
\usepackage{pifont}
\usepackage{makecell} 
\newcommand{\cmark}{\ding{51}} 
\newcommand{\xmark}{\ding{55}} 

\usepackage{newfloat}
\usepackage{listings}
\DeclareCaptionStyle{ruled}{labelfont=normalfont,labelsep=colon,strut=off} 
\floatstyle{ruled}
\newfloat{listing}{tb}{lst}{}
\floatname{listing}{Listing}
\title{CausalPulse: An Industrial-Grade Neurosymbolic Multi-Agent Copilot for \\ Causal Diagnostics in Smart Manufacturing}
\author{
    Chathurangi Shyalika\textsuperscript{\rm 1},
    Utkarshani Jaimini\textsuperscript{\rm 2},
    Cory Henson\textsuperscript{\rm 3},
    Amit Sheth\textsuperscript{\rm 1}
}
\affiliations{
     \textsuperscript{\rm 1}Artificial Intelligence Institute, University of South Carolina\\
    \textsuperscript{\rm 2}Stride Lab, University of Michigan, Dearborn
\\
    \textsuperscript{\rm 3}Bosch Center for Artificial Intelligence, Pittsburgh
}

\usepackage{bibentry}

\begin{document}

\maketitle

\begin{abstract}
Modern manufacturing environments demand real-time, trustworthy, and interpretable root-cause insights to sustain productivity and quality. Traditional analytics pipelines often treat anomaly detection, causal inference, and root-cause analysis as isolated stages, limiting scalability and explainability. In this work, we present \emph{\textbf{CausalPulse}}, an industry-grade multi-agent copilot that automates causal diagnostics in smart manufacturing. It unifies anomaly detection, causal discovery, and reasoning through a neurosymbolic architecture built on standardized agentic protocols. CausalPulse is being deployed in a Robert Bosch manufacturing plant, integrating seamlessly with existing monitoring workflows and supporting real-time operation at production scale. Evaluations on both public (\emph{Future Factories}) and proprietary (\emph{Planar Sensor Element}) datasets show high reliability, achieving overall success rates of 98.0\% and 98.73\%. Per-criterion success rates reached 98.75\% for planning and tool use, 97.3\% for self-reflection, and 99.2\% for collaboration. Runtime experiments report end-to-end latency of 50--60\,s per diagnostic workflow with near-linear scalability ($R^2{=}0.97$), confirming real-time readiness. Comparison with existing industrial copilots highlights distinct advantages in modularity, extensibility, and deployment maturity. These results demonstrate how CausalPulse’s modular, human-in-the-loop design enables reliable, interpretable, and production-ready automation for next-generation manufacturing.

\end{abstract}


\section{Introduction}

Manufacturing systems increasingly rely on dense, heterogeneous sensors and automated production lines that must maintain continuous operation under strict quality and yield constraints. Real-time detection of anomalies and accurate identification of their root causes are essential to prevent costly scrap, downtime, and latent defects. However, traditional analytic pipelines treat anomaly detection, causal discovery, and decision-making as isolated stages. Sensors feed black-box models, experts interpret alerts, and interventions occur in an ad hoc manner, leading to delayed diagnosis, limited traceability, and reduced operator trust. Reliable root cause analysis (RCA) remains difficult due to multimodal and incomplete production data \cite{kuo2021digital,pietsch2024root}, partially known and context-dependent causal relationships \cite{ko2023framework,pearl2019seven}, and the need for explainable, auditable, operator-in-the-loop systems that ensure safety and accountability \cite{sofianidis2021review,trivedi2024explainable}. Large-scale deployment requires modular, interoperable, and composable architectures to support robust integration in industrial pipelines \cite{ribeiro2021describing}.

Classical research in causal analysis and multi-agent systems provides a foundation for addressing these challenges. Structure-learning algorithms characterize conditional independence and score-based approaches to infer causal graphs from observational data \cite{spirtes2000causation,chickering2002optimal}. Multi-agent systems and service-oriented agent frameworks have demonstrated how distributed reasoning and standardized communication protocols enable modular, scalable automation in industrial settings \cite{bellifemine2000developing,wooldridge2002multi}. Recently, large language models have enabled practical methods that combine reasoning and tool use, integrating model-based planning with external tools to extend their capabilities \cite{yao2022react,schick2023toolformer}. These developments motivate the synthesis of principled causal inference, neurosymbolic domain priors, and agentic orchestration to produce interpretable, operational RCA systems. Despite promising advances, current industrial copilots and diagnostic platforms lack a unified, standards-based, multi-agent framework that (i) integrates preprocessing and multi-modal data, (ii) constrains causal analysis with domain rules, (iii) produces transparent and explainable results, (iv) supports operator-driven selection, review, and replanning at runtime, and (v) enables easy extensibility to new use cases and manufacturing tasks.

To address these gaps, we present \emph{\textbf{CausalPulse}}\footnote{Demo Video: https://www.youtube.com/watch?v=bh1XHHvqZos}, an industrial-level neurosymbolic multi-agent copilot for \emph{causal diagnostics} in smart manufacturing. Here, \emph{causal diagnostics} refers to the unified causal analysis process that encompasses both structure learning and RCA. As a copilot, CausalPulse serves as a collaborative intelligence layer that guides human–AI interaction, enabling human-in-the-loop decision-making. CausalPulse combines (1) a protocol- and framework-first agentic substrate for discoverable, composable services; (2) hybrid neurosymbolic causal discovery that blends structure learning with expert-defined rules; (3) a path-based RCA engine that scores and ranks candidate failure paths for operator usage; and (4) an adaptable workflow engine that supports operator-in-the-loop verification. We evaluated CausalPulse on two real-world use cases: a grinding process from an industrial production line and an assembly process from a research and development environment. 

\section{Related Work}
\paragraph{Agentic AI Frameworks and Orchestration Protocols.}
The rise of \emph{Agentic AI} marks a shift toward systems that couple language-model reasoning with modular orchestration and communication protocols. Frameworks such as LangGraph \cite{langgraph_2024}, AutoGen \cite{wu2024autogen}, CrewAI \cite{crewai2025}, and OpenDevin \cite{wang2024opendevin} integrate standardized interfaces for tool invocation, planning, and memory management \cite{bandi2025rise,bousetouane2025agentic}. Protocols like the Model Context Protocol (MCP) \cite{anthropic_mcp_2024}, Agent Communication Protocol \cite{agentcommprotocol2025}, Google A2A \cite{google_a2a_interoperability_2025} and Agora \cite{marro2024scalable} have formalized structured message passing and task routing, allowing autonomous agents to coordinate across services in production workflows \cite{gort2025agentedge, sawant2025unified, derouiche2025agentic}. These developments have renewed the practicality of deploying multi-agent reasoning systems in complex industrial environments.

\paragraph{Industrial and Diagnostic Copilots.}
Several domain-specific copilots adopt multi-agent and modular architectures for industrial monitoring and diagnostics. The UCSD Causal Copilot \cite{wang2025causal} integrates causal discovery and reasoning through an LLM-based planner. MATMCD \cite{shen2024exploring} and IBM AssetOpsBench \cite{patel2025assetopsbench} employ multi-agent pipelines that combine IoT data processing, anomaly detection, and maintenance scheduling. SmartPilot \cite{shyalika2025smartpilot_a, shyalika2025smartpilot_b, shyalika2025causaltrace} extends this paradigm by coupling predictive agents for process forecasting, anomaly diagnostics, and causal analysis. These systems demonstrate the feasibility of integrating multi-agent coordination and reasoning in industrial settings but remain limited in extensibility and in incorporating semantic context, domain priors, and neurosymbolic reasoning.


Despite growing interest in industrial multi-agent systems, existing copilots primarily focus on task automation, fault prediction, or isolated RCA components. They often lack an integrated framework that unifies data preprocessing, anomaly detection, causal discovery, and RCA within a cohesive agentic architecture. Current systems provide limited support for neurosymbolic reasoning and operator-driven validation. Addressing these gaps, CausalPulse introduces an interpretable, standards-based multi-agent framework that supports dynamic orchestration, knowledge-infused causal diagnostics, and operator-in-the-loop decision intelligence for smart manufacturing.

\section{CausalPulse System Overview}
\label{sec:overview}
\subsection{System Architecture and Four-layer Framework}

\begin{figure}[!ht]
\centering
\vspace{-2mm}\includegraphics[width=0.9\linewidth]{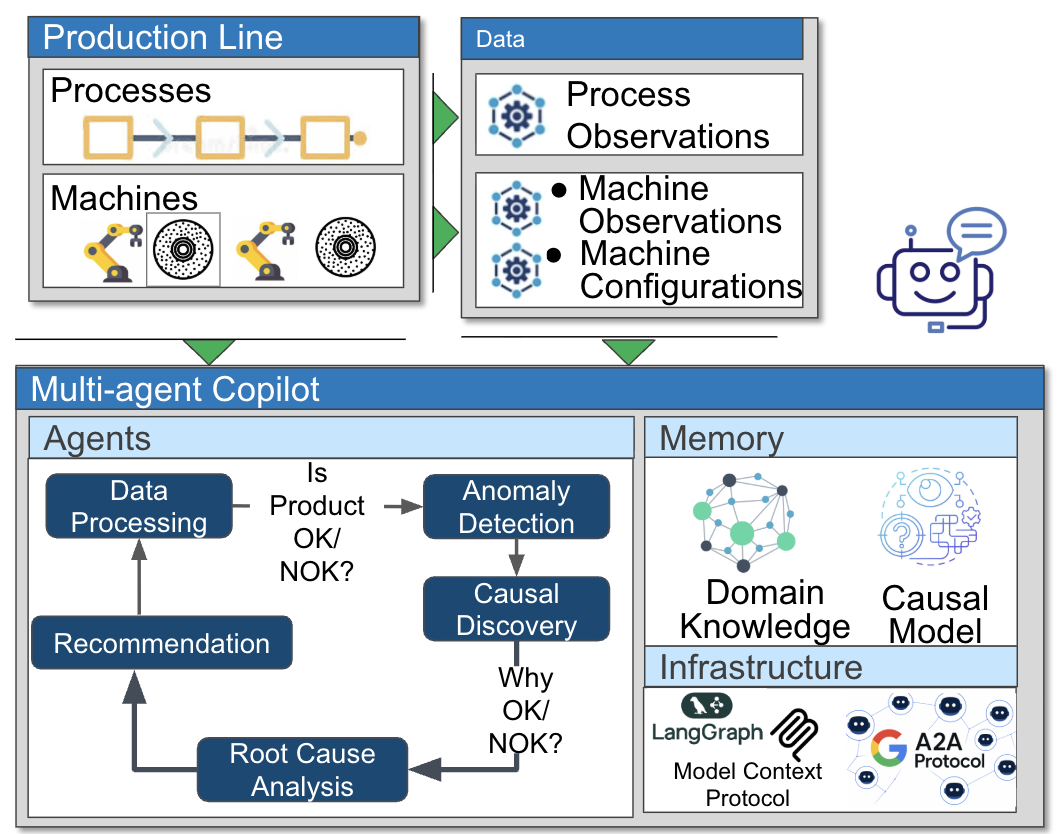}

\caption{High-level overview of CausalPulse system architecture and components.}
  \label{fig:system_architecture}

\end{figure}

\begin{figure}[!ht]
\vspace{-3 mm}\includegraphics[width=0.999\linewidth]{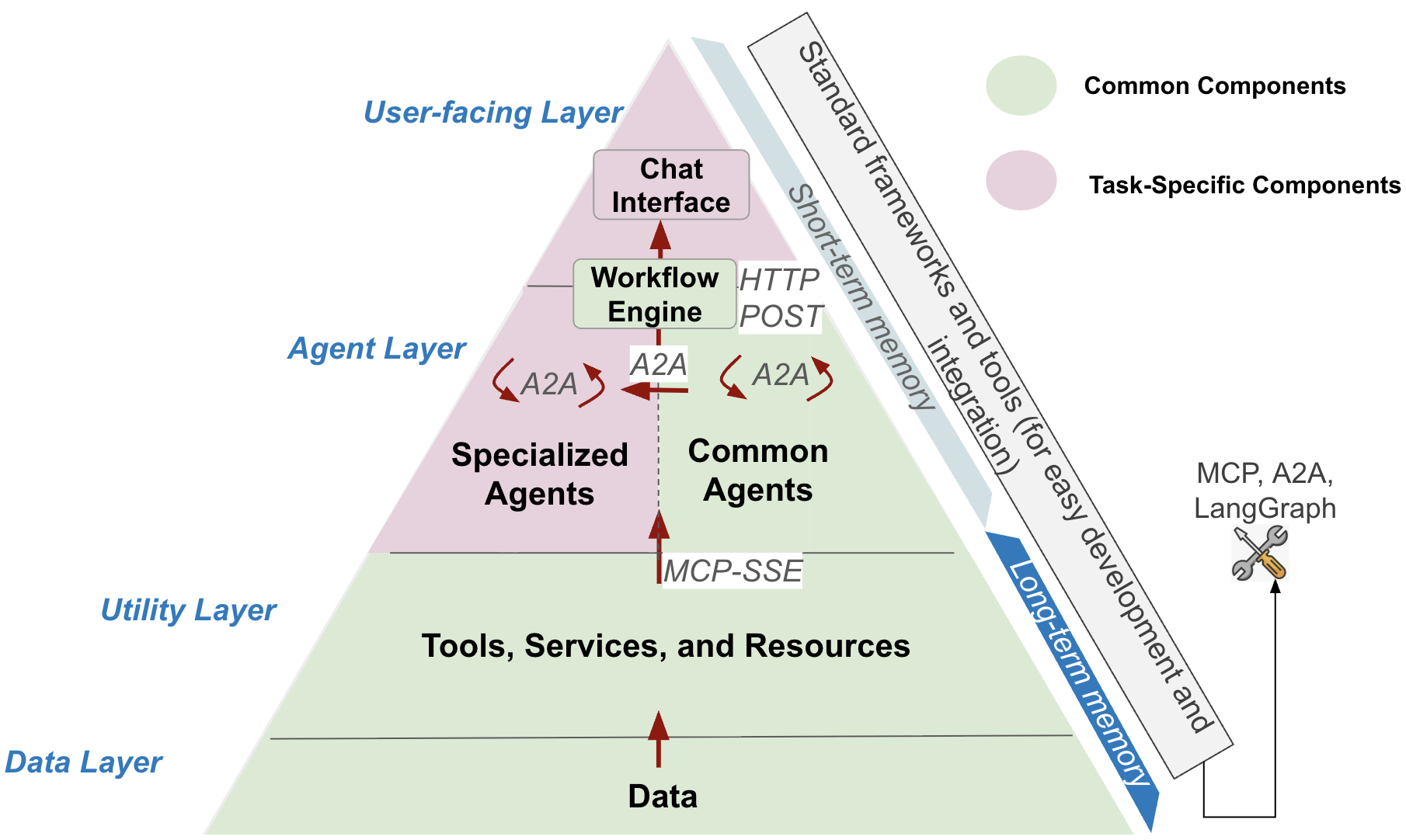}
\caption{CausalPulse four-layer framework: user-facing, agent, utility, and data, connected through standardized protocols (MCP, A2A, and LangGraph).}
  \label{fig:overall_architecture}
\end{figure}

Figures \ref{fig:system_architecture} and \ref{fig:overall_architecture} illustrate complementary perspectives of the CausalPulse platform. The high-level system architecture (Figure \ref{fig:system_architecture}) details the operational workflow, how data flows from the production line through multi-agent reasoning to generate diagnostics and recommendations. In contrast, the four-layer framework (Figure \ref{fig:overall_architecture}) formalizes the structural organization that enables this workflow, comprising the \textit{user-facing}, \textit{agent}, \textit{utility}, and \textit{data layers}. Both representations overlap conceptually; the multi-agent copilot in the system architecture maps to the agent and utility layers in the framework but serves distinct principles. The former emphasizes functional interactions, while the latter abstracts the system’s architectural hierarchy and interoperability mechanisms. The four layers of the framework are detailed below.

\subsubsection{User-Facing Layer:} Provides a browser-based user interface (UI) through which operators interact with the copilot. It enables human-in-the-loop decision-making, allowing users to view copilot outputs and refine, validate, and adjust them based on domain expertise. The LangGraph-based Workflow Engine is positioned between the UI and Agent layers. 

\subsubsection{Agent Layer:} Comprises two categories of agents: common agents, which are reusable across domains and applications, and specialized agents, which are tailored for specific causal diagnostic tasks. All agents are implemented as FastAPI microservices, described by agent cards, and coordinated through A2A communication with registry-based orchestration.

\subsubsection{Utility Layer} Provides MCP tools, services, and resources that underpin agent operations. Services integrate domain-informed services such as process ontologies, LLMs, knowledge graphs, curated technical manuals, and rule bases.  Resources are artifacts, including agent cards, preprocessed documents, stored causal graphs, RCA results, and precomputed causal knowledge.

\subsubsection{Data Layer} Includes data collected from cyber-physical systems. It encompasses sensor and process logs, camera images, event and cycle metadata, process stages, and anomalies, which may take the form of time-series, images, or textual data. These heterogeneous inputs collectively serve as the basis for higher-level agentic operations.

\subsubsection{Rationale and Design Principles.}  CausalPulse framework is conceived as a living ecosystem of \textit{data}, \textit{utilities}, \textit{agents}, and \textit{user-facing layer} to achieve modularity, explainability, and safe autonomy in complex industrial environments. Each layer in this vertical stack performs a distinct yet complementary role within the reasoning pipeline, establishing clean interfaces across human–AI–machine collaboration. The layered design bridges neural adaptability and symbolic control, enabling agents to learn from data while remaining constrained by domain knowledge. It supports plug-and-play extensibility, allowing new data sources, agents, and tools to be integrated without reconfiguring the overall workflow. The architecture ensures end-to-end transparency, allowing every decision to be traced from user query to raw data through clearly defined layer boundaries.

\subsection{Agents and Their Capabilities}
\label{sec:agents_c}

\subsubsection{Common agents. } 
The \textit{Client Process Agent} (CPA) serves as the primary orchestrator between UI, and downstream agents. It first normalizes heterogeneous client inputs into canonical payloads that downstream tools can consume. Then, based on high-level task tags (e.g., \texttt{preprocessing}, \texttt{anomaly}, \texttt{causal}), it dispatches these payloads to the appropriate tools while managing asynchronous MCP sessions and routing execution requests. 
After tool execution, the CPA postprocesses results using metadata-defined rules, optionally triggering follow-up tools to support multi-step workflows.

The \textit{Preprocessing Agent} performs deterministic data cleaning and normalization. It cleans the data by dropping empty rows and columns, standardizing types, imputing missing values, and label-encoding categorical variables into a normalized CSV with stored encoder mappings. It queries the Background Info Agent to retrieve variable-level descriptions, thereby enriching the returned summary with domain semantics and improving the interpretability of subsequent diagnostics.

The \textit{Background Info Agent} acts as a lightweight semantic lookup service that exposes curated metadata for dataset variables. Implemented as an HTTP-only microservice optimized for low-latency responses, it accepts lists of variable names and returns concise, human-readable descriptions that capture sensor semantics, measurement locations, and other domain-specific context. They help the Preprocessing and RCA agents turn numeric results into domain-aware explanations, improving operator interpretability and providing a consistent semantic layer across the diagnostic pipeline.

The \textit{Recommender Agent} provides context-aware, next-step guidance to both human operators and the automated planner as diagnostic workflows progress. It uses the current tool and outputs to suggest the next analyses, evidence to inspect, or causes to prioritize, which the user can confirm or the planner can execute automatically. Its behavior is governed by deterministic, rule-based heuristics rather than opaque learned policies, ensuring predictable responses and low inference latency. Recommendations are exposed via the MCP SSE interface, enabling seamless discovery and consumption by the CPA and other A2A callers.
\\
\vspace{-3mm}
\subsubsection{Specialized agents. } 
The \textit{Anomaly Detection Agent} encapsulates a multi-strategy pipeline for detecting abnormal sensor and process behavior. 
Based on the query content and available data modalities, the agent selects among several anomaly detection backends: a cross-modal fusion model combining time-series and image-based signals \cite{shyalika2025nsfmap}, deterministic threshold rules derived from domain-specific measurement limits, and an isolation-forest model for unsupervised detection on tabular features. Its outputs comprise categorical labels (e.g., \texttt{Normal} vs.\ specific anomaly types), anomaly scores, and explicit identification of features that violate expected bounds, and traceability metadata. When an anomaly is detected, the agent automatically constructs a structured payload summarizing the event and triggers the RCA Agent via A2A communication.



\begin{figure}[t]
\centering
\vspace{-2mm}
\begin{tikzpicture}[
    >=Stealth,
    node distance=1.6cm,
    control/.style={rectangle, draw, rounded corners, minimum width=1.9cm,
                    minimum height=0.7cm, align=center, fill=blue!10},
    obs/.style={circle, draw, minimum size=0.9cm, align=center, fill=orange!10},
    legend/.style={rectangle, draw, rounded corners, minimum width=1.6cm,
                   minimum height=0.6cm, align=center},
    valid/.style={->, thick},
    invalid/.style={->, thick, dashed, red}
]

\node (seqlabel) {\small Event sequence: $G_4 \prec G_{10} \prec \dots$};
\vspace{-4mm}
\node[control, below=1.0cm of seqlabel] (c4) {Control\\$G_4$};
\node[obs, below=1.4cm of c4] (o4) {Obs$_4$};

\node[control, right=2.8cm of c4] (c10) {Control\\$G_{10}$};
\node[obs, below=1.4cm of c10] (o10) {Obs$_{10}$};

\draw[valid] (c4) -- node[left=2pt] {\scriptsize R1,R5} (o4);
\draw[valid] (c4) -- node[above=2pt, sloped, midway] {\scriptsize R2} (o10);
\draw[valid] (o4) -- node[below=2pt] {\scriptsize R3} (o10);

\draw[invalid] (c4) .. controls +(1.0,0.9) and +(-1.0,0.9) ..
  node[above=2pt, sloped, midway] {\scriptsize R1} (c10);

\draw[invalid] (c10) -- node[above=2pt, sloped, midway] {\scriptsize R2} (o4);

\draw[invalid] (o10) -- node[below=2pt, sloped, pos=0.25] {\scriptsize R4} (o4);

\draw[invalid] (o4) .. controls +(-1.0,-0.9) and +(1.0,-0.9) ..
  node[below=0pt, sloped, midway] {\scriptsize R3} (c10);
\vspace{-10mm}
\node[legend, below=0.5cm of o4, xshift=-0.5cm] (lg1) {Control parameter};
\node[control, right=0.3cm of lg1] {};

\node[legend, below=0.25cm of lg1, xshift=0cm] (lg2) {Observation parameter};
\node[obs, right=0.3cm of lg2] {};

\node[legend, below=0.2cm of lg2, xshift=0cm] (lg3) {Valid causal edge};
\draw[valid] ([xshift=0.3cm]lg3.east) -- ++(1.0,0);

\node[legend, below=0.2cm of lg3, xshift=0cm] (lg4) {Forbidden causal edge};
\draw[invalid] ([xshift=0.3cm]lg4.east) -- ++(1.0,0);

\end{tikzpicture}
\caption{Illustration of domain-specific rules for a grinding process. 
Control parameters (rectangles) may cause observation parameters (circles) at the same or later grinding stage (Rules R1, R2, R5),
observation parameters may only cause later observations (Rules R3, R4),
and edges violating type or temporal order are explicitly disallowed (dashed red arrows).}
\label{fig:domain_rules}
\vspace{-5mm}
\end{figure}
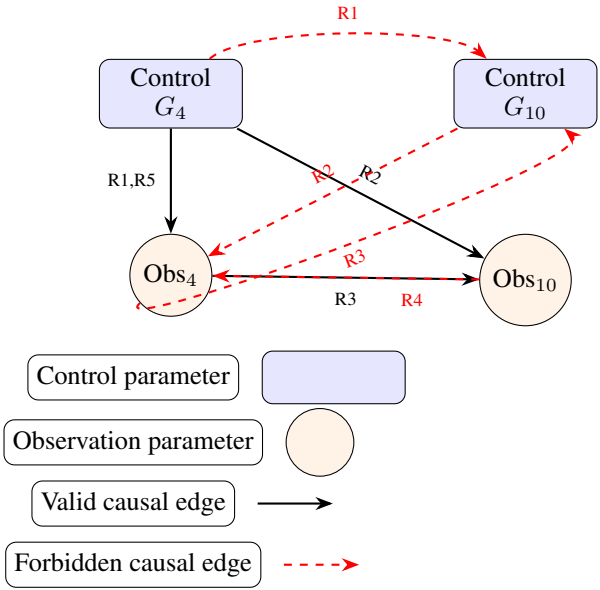

The \textit{Causal Discovery Agent} induces candidate causal structures from preprocessed data in a schema-aware, knowledge-infused manner. 
Based on the uploaded data, it selects an appropriate discovery routine—using Peter–Clark (PC) for gripper-sensor datasets and Greedy Equivalence Search (GES) for process-variable datasets to construct causal graphs. During structure learning, both algorithms are instantiated via the \texttt{causal-learn} \cite{causallearn_2024} and \texttt{pgmpy} \cite{pgmpy_2024} and are constrained by domain-specific rules on control/observation types and process order (Figure \ref{fig:domain_rules}), ensuring that only physically plausible edges are considered. The resulting directed graphs and edge-level statistics, filtered by these domain constraints, can further be inspected and refined by factory-floor experts.

\begin{figure*}[!ht]
\centering
\vspace{-2mm}
\includegraphics[width=0.8\linewidth]{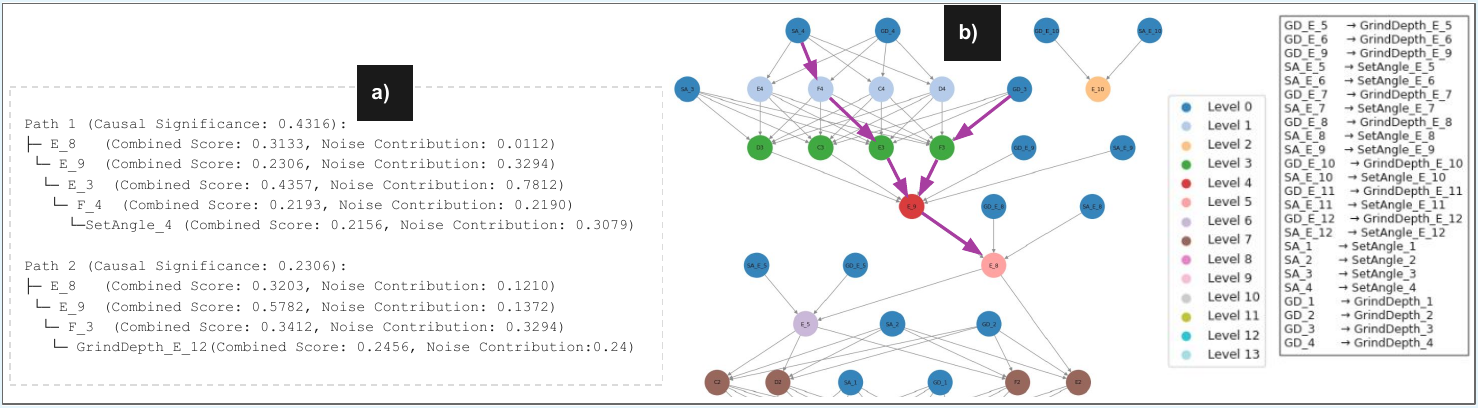}
\vspace{-2mm}
\caption{Root cause analysis results: 
(a) ranked root-cause paths ordered by causal significance for the anomaly in E\_8 in the PSE dataset (refer to "Datasets Descriptions" Section), and 
(b) visual representation of the corresponding causal graph, highlighting identified root causes and propagation paths. The legend on the right lists variable names and their hierarchical levels, where levels represent the sequential order of grinding steps in the PSE use case.}
\label{fig:RCA}
\vspace{-4mm}
\end{figure*}

The \textit{RCA Agent} is an MCP- and HTTP-capable component that performs causal attribution on anomalous observations. It accepts either in-memory records (reusing precomputed RCA results if available), an uploaded dataset, or a file path, and returns structured, ranked root-cause explanations suitable for operator interpretation and downstream tooling. The agent wraps the \emph{ProRCA} analysis pipeline—a domain-informed SCM builder and path-based root-cause scorer~\cite{dawoud2025prorca}. It outputs ranked candidate causes, per-node scores, and enumerated causal paths. Scoring combines structural and noise-based measures to balance causal plausibility with observed perturbation strength, enhancing operator trust and guiding follow-up diagnostics (Figure~\ref{fig:RCA}).


\subsection{Agentic Task Flow Orchestration and Workflows}

\begin{figure*}[!ht]
\centering
\vspace{-3mm}\includegraphics[width=0.8\linewidth, height=8cm]{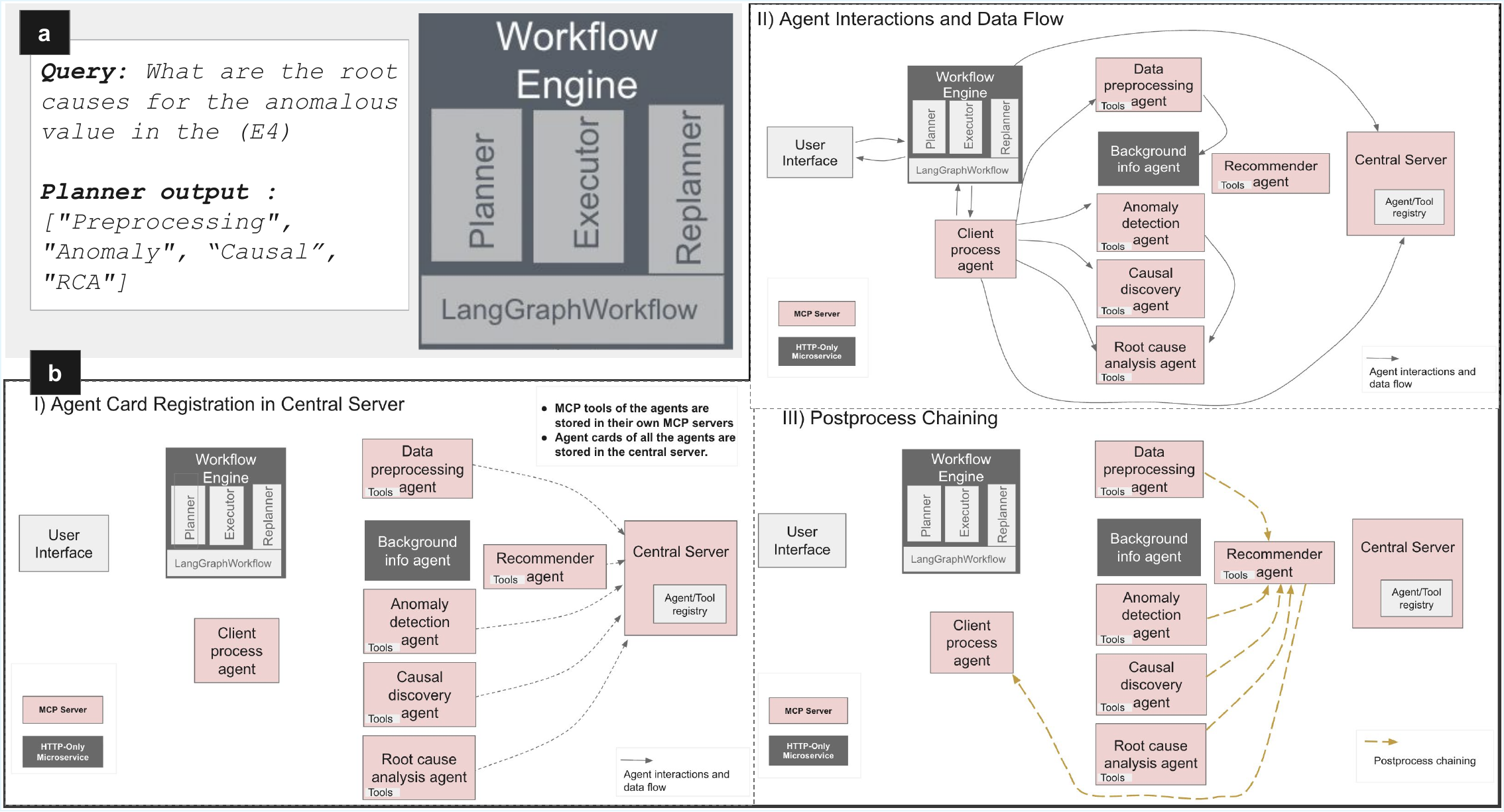}
  \vspace{-2mm}
\caption{Agentic task orchestration and workflow execution in CausalPulse: 
(a) Query interpretation and planning by the LangGraph-based workflow engine, 
(b-I) Agent card registration and central server coordination, 
(b-II) Agent interactions and data flow during execution, and 
(b-III) Postprocess chaining with the Recommender Agent.}
  \label{fig:task_flow}
\end{figure*}

The workflow of CausalPulse includes three stages.\\
\textbf{Stage 1: Agent card registration (Figure \ref{fig:task_flow}:b-I):} Every agent registers its agent card on the central server. This registration process allows the central server to maintain a live directory of all available agents, enabling dynamic discovery and orchestration across the system \\
\textbf{Stage 2: Agent interactions and data flow (Figure \ref{fig:task_flow}:b-II):} Once the agents are registered, the workflow begins when a user query is submitted through the interface. Within the workflow engine (Figure~\ref{fig:task_flow}a), the \emph{planner node} receives the user’s query together with the current workflow state. It queries an LLM (\texttt{llama-3.1-8b-instant}) to determine the minimal sequence of agents required to address the query, such as preprocessing, anomaly detection, causal discovery, and RCA. This process constitutes dynamic task decomposition, ensuring that redundant steps are avoided, such as reapplying preprocessing when it has already been completed. The node then updates the system state with the planned steps and returns this structured plan to the workflow.

The \emph{executor node} is responsible for carrying out the planned steps. For every stage the executor sends a structured request to the CPA. CPA dispatches the request to the correct remote agent via the MCP protocol and SSE connection. It returns the agent’s output back to the executor, which merges results into the shared system state before moving on to the next stage. For efficiency, previously completed stages are skipped, such as preprocessing when it has already been performed. The executor also ensures proper data handling by forwarding file names and data paths, which is particularly important for consecutive agents. After execution, results are appended into the appropriate state slot (e.g., \emph{state["anomaly"]}, \emph{state["RCA"]}), and the workflow continues to the next step.

The \emph{replanner node} is invoked after each execution to evaluate if the plan requires modification. It queries the LLM to assess whether additional stages are needed based on the updated state and prior results. For instance, if anomaly detection indicates a \emph{"normal"} status, the system excludes unnecessary stages such as RCA or causal discovery. The replanner then returns an updated list of remaining steps, ensuring that the workflow adapts dynamically to the context.

The LangGraph workflow construction specifies the orchestration of these nodes. The workflow begins with the planner, proceeds to the executor, and transitions to the replanner as long as steps remain. If additional steps are identified, the cycle continues with replanning and execution. If no steps remain, the workflow terminates. 

\textbf{Stage 3: Postprocess chaining (Figure \ref{fig:task_flow}:b-III):} After the primary agents complete their tasks, the Recommender Agent is automatically activated. It reviews the aggregated outputs and the system’s current state, then suggests the next best actions. Upon receiving these recommendations, the CPA validates and merges them into the shared workflow state, exposes the suggested actions through the UI, and evaluates any postprocess rules defined in the agent cards. When auto-chaining is permitted, the CPA resolves mapped inputs and triggers the corresponding downstream agents via standardized MCP calls. 
Returned results are integrated into the global state and recorded as entries for traceability. If new outputs alter the workflow context, the CPA invokes the replanner node to dynamically update the execution plan, ensuring adaptive, continuous reasoning.

\vspace{-1mm}



\subsection{Knowledge Representation and Semantics}
\label{subsec:knowledge_semantics}

CausalPulse grounds its agentic workflows in semantic artifacts capturing variable meaning, process structure, and causal constraints. 
\paragraph{Variable-Level Semantics.}
CausalPulse maintains curated metadata for each sensor or process variable, including textual descriptions, units, physical location, subsystem membership, and high-level type (e.g., control vs.\ observation, quality vs.\ condition). 
This metadata is stored in as JSON and accessed through the Background Info Agent, which exposes a simple lookup API from variable names to human-readable descriptions.
The Preprocessing Agent uses this catalog to enrich preprocessing summaries with semantic context, while the RCA and Causal Discovery Agents incorporate it into their explanations.

\paragraph{Process and Asset Structure.}
Process-level semantics are encoded as directed graphs covering process stages and asset hierarchies (machine $\rightarrow$ subsystem $\rightarrow$ sensor) defining the temporal sequence and stage-specific control/observation variables. 
They help agents filter relevant variables, group them by stage, and give the planner a coarse sense of earlier vs. later process regions when building workflows.

\paragraph{Causal and Operational Rules.}
Domain experts define rule sets that restrict which causal relations between variables are admissible.
Example rules for PSE grinding process are visualized in Figure \ref{fig:domain_rules}.
These rules are encoded in a compact declarative schema over variable types and stage indices and enforced as hard constraints during causal structure learning. The Causal Discovery Agent uses them to filter candidate edges from PC and GES, retaining only rule-consistent relationships, which reduces spurious correlations and yields causal graphs that are physically plausible and aligned with operators’ mental models of the line.

\paragraph{Generalization to New Lines.}
The underlying representation of rule sets and process graphs in this work is designed to be portable.
Onboarding a new manufacturing line requires (i) ingesting its process topology and stage ordering into the process graph schema, (ii) populating the variable catalog with descriptions and type labels, and (iii) eliciting a small number of control/observation and order rules from domain experts.
Once registered, these artifacts are consumable by agents and planners without architectural changes, allowing semantic grounding and knowledge-infused causal discovery to carry over across lines and sites.

\section{Results and Evaluation}
\label{sec:evaluation}
\subsection{Datasets Descriptions}
\subsubsection{Planar Sensor Element (PSE) dataset.}
\label{sec:pse}
The PSE dataset is collected from Bosch’s Anderson Plant (AdP) in South Carolina, USA, where planar oxygen sensors are manufactured. It captures the grinding process, a precision operation critical to product quality and yield. This process spans 18 robotic stations with dual-arm configurations performing 12 sequential grind types ($G_1$–$G_{12}$). Each grind is defined by parameters such as angle, depth, duration, and speed, with repetitions possible upon anomaly detection or operator intervention. For each grind, distances (mm) and angles (degrees) are measured at multiple surface positions, producing 26 variables through optical inspection and machine sensors. Each feature includes defined tolerance limits that adjust dynamically for regrinds to avoid over-polishing or material stress. 

\subsubsection{Future Factories (FF) Dataset.}
We use the publicly available manufacturing dataset generated by the Future Factories (FF) Lab at the McNair Aerospace Research Center, University of South Carolina \cite{harik2024analog}. This dataset captures synchronized sensor and image data from a prototype rocket assembly pipeline designed to emulate industrial-grade manufacturing processes. Each cycle is segmented into 21 distinct operational states. The dataset includes
time-series measurements from potentiometers, load cells,
drive temperatures, robot kinematics and synchronized image data captured from two cameras.  These anomalies indicate missing rocket components and include six types, such as \texttt{NoNose}, \texttt{NoBody2}, and combined cases like \texttt{NoBody2,NoBody1}. 

CausalPulse delivers interpretable results, reducing manual adjustments in high-cost manufacturing, with a 4\% scrap rate ($\approx$\$ 15M annually) in the PSE dataset and a $\approx$\ 40\% anomaly rate in the FF dataset.




\subsection{Evaluation Setup}

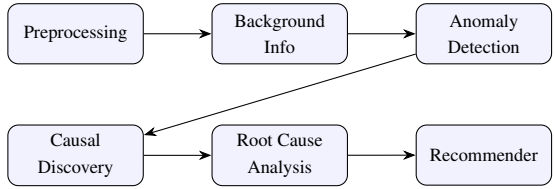
\begin{figure}[t]
\centering
\begin{tikzpicture}[
    >=Stealth,
    node distance=0.8cm and 0.9cm,
    block/.style={rectangle, draw, rounded corners,
                  minimum width=1.8cm, minimum height=0.8cm,
                  align=center, fill=blue!5, font=\scriptsize}
]

\node[block] (P)  {Preprocessing};
\node[block, right=of P]  (BI)  {Background\\Info};
\node[block, right=of BI] (AD)  {Anomaly\\Detection};

\node[block, below=of P]  (CD)  {Causal\\Discovery};
\node[block, below=of BI] (RCA) {Root Cause\\Analysis};
\node[block, below=of AD] (R)   {Recommender};

\draw[->] (P)  -- (BI);
\draw[->] (BI) -- (AD);
\draw[->] (AD) -- (CD);
\draw[->] (CD) -- (RCA);
\draw[->] (RCA) -- (R);

\end{tikzpicture}
\caption{Canonical agent pipeline in CausalPulse. Predefined workflows correspond to subpaths of this pipeline with optional reuse of cached intermediate results.}
\label{fig:canonical_workflow}
\vspace{-4mm}
\end{figure}

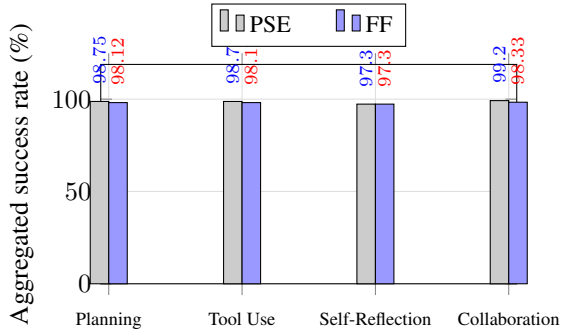
\begin{figure}[!ht]
\centering
\begin{tikzpicture}
\begin{axis}[
    width=0.4\textwidth,
    height=4.5cm,
    ybar=0pt,
    bar width=7pt,
    enlarge x limits=0.02,
    ymin=0, ymax=110,
    ylabel={Aggregated success rate (\%)},
    symbolic x coords={Planning,Tool Use,Self-Reflection,Collaboration},
    xtick=data,
    xticklabel style={font=\scriptsize, yshift=-3pt},
    legend columns=2,
    legend style={
        /tikz/every even column/.append style={column sep=1.5em},
        at={(0.5,1.08)},
        anchor=south,
        draw=black,
        fill=white,
        font=\footnotesize
    },
    nodes near coords,
    every node near coord/.append style={
        font=\scriptsize,
        yshift=3pt,
        anchor=mid west,
        rotate=90,
    },
    point meta=y,
    enlarge y limits={upper,value=0.08},
    clip=false,
    grid=major,
    major grid style={gray!25},
]
\addplot+[draw=black, fill=gray!40] coordinates {
    (Planning,98.75) (Tool Use,98.75) (Self-Reflection,97.3) (Collaboration,99.2)
};
\addplot+[draw=black, fill=blue!40] coordinates {
    (Planning,98.12) (Tool Use,98.12) (Self-Reflection,97.3) (Collaboration,98.33)
};
\legend{PSE, FF}
\end{axis}
\end{tikzpicture}
\caption{Aggregated success rate per agentic criterion for PSE and FF datasets.}
\label{fig:agentic_patterns_compare}
\vspace{-2mm}
\end{figure}

We evaluated CausalPulse’s core agentic capabilities using predefined workflows for diagnostic queries (Figure \ref{fig:canonical_workflow}, Table~\ref{tab:workflows}). The responses were analyzed to determine the number of expected and predicted reasoning steps and agents, along with successful and unsuccessful attempts. We computed three quantitative metrics, \textit{Criterion success (\%)}, \textit{Aggregate criterion success (\%) per stage}, and the \textit{Aggregate success rate per agentic criterion}. The \textit{Criterion success (\%)} represents the proportion of successful executions for each agentic criterion, such as planning, tool use, self-reflection, and multi-agent collaboration, calculated as the ratio of successful to total queries. The \textit{Aggregate criterion success (\%) per stage} summarizes the overall success rate of a criterion type across all diagnostic stages within each dataset, computed as the ratio of total successful attempts to total queries for that criterion within a workflow stage. \textit{Aggregate success rate per agentic criterion} computes the combined reliability of each criterion type aggregated across all workflow stages. A high \textit{Criterion success (\%)} indicates close alignment between CausalPulse’s predicted reasoning trajectory and the expected expert-designed plan, while high \textit{Aggregate criterion success (\%)} and \textit{Aggregate success rate per agentic criterion} reflect robust and reliable multi-agent orchestration across workflows (Table~\ref{tab:recomputed_metrics} and Figure~\ref{fig:agentic_patterns_compare}).

\begin{table}[t]
\centering
\setlength{\tabcolsep}{3pt}
\renewcommand{\arraystretch}{1.05}
\begin{tabular}{@{}cl@{}}
\toprule
\textbf{ID} & \textbf{Workflow path (agents in order)} \\
\midrule
W1 & Preprocessing $\rightarrow$ Recommender \\
W2 & Preprocessing $\rightarrow$ Background Info $\rightarrow$ Recommender \\
W3 & Background Info $\rightarrow$ Recommender \\
W4 & Preprocessing$^{*}$ $\rightarrow$ Anomaly Detection \\ & $\rightarrow$ Recommender \\
W5 & Preprocessing$^{*}$ $\rightarrow$ Background Info $\rightarrow$ \\ & Causal Discovery $\rightarrow$ Recommender \\
W6 & Preprocessing$^{*}$ $\rightarrow$ Anomaly Detection $\rightarrow$ \\ & Causal Discovery$^{*}$  $\rightarrow$ RCA $\rightarrow$ Recommender \\
W7 & Preprocessing$^{*}$ $\rightarrow$ Anomaly Detection $\rightarrow$ \\ & Causal Discovery  $\rightarrow$  Recommender \\
W8 & Preprocessing $\rightarrow$ Anomaly Detection $\rightarrow$ Background \\ & Info  $\rightarrow$ Causal Discovery $\rightarrow$ RCA $\rightarrow$ Recommender \\
W9 & Preprocessing$^{*}$ $\rightarrow$ Anomaly Detection $\rightarrow$ \\ & Causal Discovery$^{*}$ $\rightarrow$ RCA $\rightarrow$ Recommender \\
\bottomrule
\end{tabular}
\caption{Predefined workflows in CausalPulse as paths through the canonical agent pipeline in Figure \ref{fig:canonical_workflow}.  
Steps marked $^{*}$ may reuse cached results if available.}
\label{tab:workflows}
\vspace{-4mm}
\end{table}

\label{subsec:agent_capabilities}

\textbf{Evaluating Planning Capability:} We assessed the ability of CausalPulse’s workflow engine to correctly decompose user intents into coherent subtasks following canonical workflow patterns. Across both datasets, the system achieved a planning success rate  of 98.75\% (PSE) and 98.12\% (FF).
The planner generated the correct sequence of reasoning steps and agents, with minor discrepancies in cases where intermediate causal diagnostics steps were omitted due to incomplete contextual cues.

\textbf{Evaluating Tool Use:} Tool invocation success rate was evaluated by comparing the tools executed per query against ground-truth workflows.
CausalPulse achieved tool-use success rates of 98.75\% (PSE) and 98.1\% (FF), reflecting reliable orchestration of MCP tools through LangGraph and A2A protocols.

\textbf{Evaluating Self-Reflection:}
Self-reflection is evaluated as the system’s ability to introspect on prior actions and reuse intermediate results instead of recomputing them. 
CausalPulse exhibited three reflective behaviors:
(i) dynamically pruning unnecessary tasks when upstream outputs render them obsolete (e.g., skipping RCA when the anomaly detector returns ``normal''), 
(ii) avoiding redundant preprocessing and causal discovery steps when prior results are available, and 
(iii) reusing cached intermediate results for recurrent subqueries. 
Empirical analysis shows a reflective success rate of 97.3\% for both datasets, while redundant computation was reduced by approximately 22\% across repeated workflow executions. 

\textbf{Evaluating Multi-Agent Collaboration:} Multi-agent collaboration was assessed based on the system’s ability to invoke the correct subset of agents and exchange intermediate results through standardized A2A communication.
The collaboration success rate reached 99.2\% (PSE) and 98.3\% (FF), demonstrating robust coordination and information sharing across agents.
Occasional collaboration lapses were limited to ambiguous user intents where the recommender agent prematurely finalized responses before RCA completion. 

CausalPulse achieved high criterion-level performance across both datasets, with per-criterion success rates ranging from 95–100\%. The aggregate criterion success reached 98.73\% for PSE and 98.0\% for FF, indicating stable and reliable multi-agent execution across workflow stages (Table \ref{tab:recomputed_metrics}).

\begin{table*}[!ht]
\centering
\caption{Criterion success and aggregated criterion success per stage across datasets, workflow stages, and agentic criteria}
\label{tab:recomputed_metrics}
\renewcommand{\arraystretch}{1.05}
\setlength{\tabcolsep}{4pt}
\resizebox{\textwidth}{!}{%
\begin{tabular}{lllcccccccc}
\toprule
\textbf{Dataset} & \textbf{Stage} & \textbf{Criteria} & 
\textbf{No.of Queries} & 
\textbf{Expected (steps/agents)} & 
\textbf{Predicted (steps/agents)} &
\textbf{Successful attempts} & 
\textbf{Failed attempts} & 
\textbf{Criterion success (\%)} & 
\textbf{Aggregate criterion success (\%)} \\
\midrule
\multirow{16}{*}{PSE}
 & \multirow{4}{*}{Preprocessing}
   & Planning             & 20 & 3 & 3 & 20 & 0 & 100.0 & 98.73 \\
 &  & Tool use            & 20 & 2 & 2 & 20 & 0 & 100.0 & 98.73 \\
 &  & Self-reflection     & 20 & 3 & 2 & 19 & 1 & 95.0 & 98.73 \\
 &  & Multi-agent collab. & 20 & 3 & 3 & 19 & 1 & 95.0 & 98.73 \\
\cmidrule(lr){2-10}
 & \multirow{4}{*}{Anomaly detection}
   & Planning             & 50 & 3 & 3 & 50 & 0 & 100.0 & 98.73 \\
 &  & Tool use            & 50 & 3 & 3 & 50 & 0 & 100.0 & 98.73 \\
 &  & Self-reflection     & 20 & 3 & 2 & 18 & 2 & 90.0 & 98.73 \\
 &  & Multi-agent collab. & 20 & 3 & 3 & 20 & 0 & 100.0 & 98.73 \\
\cmidrule(lr){2-10}
 & \multirow{4}{*}{Causal discovery}
   & Planning             & 50 & 3 & 3 & 49 & 1 & 98.0 & 98.73 \\
 &  & Tool use            & 50 & 3 & 3 & 49 & 1 & 98.0 & 98.73 \\
 &  & Self-reflection     & 30 & 3 & 3 & 30 & 0 & 100.0 & 98.73 \\
 &  & Multi-agent collab. & 30 & 3 & 3 & 30 & 0 & 100.0 & 98.73 \\
\cmidrule(lr){2-10}
 & \multirow{4}{*}{Root Cause Analysis}
   & Planning             & 40 & 5 & 5 & 39 & 1 & 97.5 & 98.73 \\
 &  & Tool use            & 40 & 5 & 5 & 39 & 1 & 97.5 & 98.73 \\
 &  & Self-reflection     & 40 & 4 & 3 & 40 & 0 & 100.0 & 98.73 \\
 &  & Multi-agent collab. & 50 & 5 & 5 & 50 & 0 & 100.0 & 98.73 \\
\midrule
\multirow{16}{*}{FF}
 & \multirow{4}{*}{Preprocessing}
   & Planning             & 20 & 3 & 3 & 19 & 1 & 95.0 & 98.00 \\
 &  & Tool use            & 20 & 2 & 2 & 19 & 1 & 95.0 & 98.00 \\
 &  & Self-reflection     & 20 & 3 & 2 & 20 & 0 & 100.0 & 98.00 \\
 &  & Multi-agent collab. & 20 & 3 & 3 & 19 & 1 & 95.0 & 98.00 \\
\cmidrule(lr){2-10}
 & \multirow{4}{*}{Anomaly detection}
   & Planning             & 50 & 3 & 3 & 50 & 0 & 100.0 & 98.00 \\
 &  & Tool use            & 50 & 3 & 3 & 50 & 0 & 100.0 & 98.00 \\
 &  & Self-reflection     & 20 & 3 & 2 & 18 & 2 & 90.0 & 98.00 \\
 &  & Multi-agent collab. & 20 & 3 & 3 & 20 & 0 & 100.0 & 98.00 \\
\cmidrule(lr){2-10}
 & \multirow{4}{*}{Causal discovery}
   & Planning             & 50 & 3 & 3 & 49 & 1 & 98.0 & 98.00 \\
 &  & Tool use            & 50 & 3 & 3 & 49 & 1 & 98.0 & 98.00 \\
 &  & Self-reflection     & 30 & 3 & 3 & 29 & 1 & 96.7 & 98.00 \\
 &  & Multi-agent collab. & 30 & 3 & 3 & 29 & 1 & 96.7 & 98.00 \\
\cmidrule(lr){2-10}
 & \multirow{4}{*}{Root Cause Analysis}
   & Planning             & 40 & 5 & 5 & 39 & 1 & 97.5 & 98.00 \\
 &  & Tool use            & 40 & 5 & 5 & 39 & 1 & 97.5 & 98.00 \\
 &  & Self-reflection     & 40 & 4 & 3 & 40 & 0 & 100.0 & 98.00 \\
 &  & Multi-agent collab. & 50 & 5 & 5 & 50 & 0 & 100.0 & 98.00 \\
\bottomrule
\vspace{-10mm}
\end{tabular}
\vspace{-6mm}
}
\end{table*}

\textbf{Runtime and Scalability Analysis: }
Runtime experiments were performed for the same set of diagnostic queries on a MacBook Pro (Apple M1 Pro, 8-core CPU, 16 GB RAM, macOS Monterey 12.7.5). On the PSE dataset ($M\!=\!5$ variables, $N\!\approx\!1.0$\,M samples), data preprocessing averaged 15s, anomaly detection 5s, causal discovery with the knowledge-infused GES algorithm 12s, and RCA $\approx$\,21s.  
End-to-end latency from query submission to RCA was $\approx$\,53s per request.  
On the FF dataset ($M\!=\!5$ variables, $N\!\approx\!1.0$\,M samples), preprocessing averaged 16s, anomaly detection 6s, causal discovery 15s, and RCA $\approx$\,23s, with a total latency of $\approx$\,60s.  
End-to-end latency represents total wall-clock time across all pipeline stages.  
To assess scalability, $M$ was increased from 5 to the full dimensionality of each dataset.  
Runtime grew approximately linearly with $M$ (coefficient of determination $R^2\!=\!0.97$), confirming near-linear scaling of the microservice architecture and asynchronous task queue.

\textbf{Root-Cause Attribution Verification and Quantitative Evaluation}
To verify the quality of PRORCA's root-cause attribution, we manually inspected anomalous events and annotated the set of ``immediate variables changed'' at the anomaly timestamp as ground-truth causes. For each event, we compared the ranked variable list produced by PRORCA against these annotations using two families of metrics: (i) rank-based \textit{hits@k}, which checks whether any true cause appears within the top-$k$ positions of the PRORCA ranking, and (ii) set-based precision, recall, and F1 between the predicted causes and the annotated set. Across all anomaly instances in the PSE dataset, PRORCA attains $\mathrm{Hits@1} = 0.44$ and $\mathrm{Hits@2} = 0.65$, rising to $0.73$ for $k \geq 3$. It achieves an average precision of 0.46, recall of 0.40, and F1 of 0.39 aggregating all anomalies in PSE dataset. Representative qualitative outcomes for three anomalies are shown in Table~\ref{tab:rca-examples}. PRORCA frequently surfaces at least one correct causal variable near the top of its ranking, but still misses a non-trivial fraction of annotated causes, leaving room for improving recall while preserving its relatively sharp rankings.

\begin{table}[!ht]
\centering
\setlength{\tabcolsep}{4pt} 
\resizebox{\linewidth}{!}{%
\begin{tabular}{l p{3cm} p{3cm} p{2cm} ccc}
\toprule
Anom. & Ground-truth causes & PRORCA top predictions & Hits@k &
Prec. & Rec. & F1 \\
\midrule
F4 &
\{\mbox{SetAngle\_3}, \mbox{GrindDepth\_3}, \mbox{GrindDepth\_4}\} &
\{\mbox{SetAngle\_4}, \mbox{GrindDepth\_4}\} &
hits@1 = 0,\; hits@2 = 1 &
0.50 & 0.33 & 0.40 \\
\midrule
E12 &
\{\mbox{GrindDepth\_E\_9}, \mbox{GrindDepth\_E\_12}\} &
\{\mbox{SetAngle\_3}, \mbox{GrindDepth\_E\_12}, \mbox{GrindDepth\_E\_9},
 \mbox{SetAngle\_4}, \mbox{SetAngle\_2}\} &
hits@1 = 0,\; hits@2 = 1,\; hits@3 = 1,\newline
hits@4 = 1,\; hits@5 = 1 &
0.40 & 1.00 & 0.57 \\
\midrule
E3 &
\{\mbox{SetAngle\_3}, \mbox{GrindDepth\_3}\} &
\{\mbox{GrindDepth\_3}, \mbox{SetAngle\_3}, \mbox{SetAngle\_4},
 \mbox{GrindDepth\_4}\} &
hits@1 = 1,\; hits@2 = 1,\; hits@3 = 1,\newline
hits@4 = 1 &
0.50 & 1.00 & 0.67 \\
\bottomrule
\end{tabular}
}
\caption{Representative PRORCA root-cause attribution outcomes on the PSE dataset, illustrating (top) a partial match where only one of three ground-truth causes appears in the top-2 positions, (middle) a high-recall but noisy case where all ground-truth causes are recovered along with several false positives, and (bottom) a strong match where both ground-truth causes are ranked in the top-2 positions.}
\label{tab:rca-examples}
\vspace{-8mm}
\end{table}

\begin{table}[!ht]
\centering
\scriptsize
\setlength{\tabcolsep}{2pt}   
\renewcommand{\arraystretch}{1.0}

\begin{tabular}{@{}l*{5}{@{\hspace{3pt}}c}@{}}
\toprule
 & \textbf{UCSD} & \textbf{MATMCD} & \textbf{AssetOps} & \textbf{SmartPilot} & \textbf{CausalPulse} \\
\midrule
1. System type
 & \shortstack{Single\\agent}
 & \shortstack{Multi-agent\\discovery}
 & \shortstack{Multi-agent\\simulation}
 & \shortstack{Multi-agent\\copilot}
 & \textbf{\shortstack{Industry-grade\\multi-agent\\copilot}} \\
\midrule
2. End-to-end \\ causal diagnostics \\pipeline
 & \xmark & \xmark & \xmark & \xmark & \cmark \\
\midrule
3. Dynamic \\multi-agent\\orchestration \\(A2A,replanning)
 & \xmark & \xmark & \xmark & \xmark & \cmark \\
\midrule
4. Knowledge-\\grounded /\\neurosymbolic \\reasoning
 & \xmark & \xmark & \xmark & \cmark & \cmark \\
\midrule
5. Live / in-factory\\deployment underway
 & \xmark & \xmark & \xmark & \cmark & \cmark \\
\midrule
6. Operator-facing\\dashboard
 & \xmark & \xmark & \xmark & \cmark & \cmark \\
\bottomrule
\end{tabular}

\caption{Summary comparison of CausalPulse with representative industrial copilots.}
\label{tab:summary_comparison}
\vspace{-4mm}
\end{table}

\textbf{Qualitative Comparison with Existing Copilots:}
To situate CausalPulse within the landscape of industrial copilots and multi-agent systems, we qualitatively compare it against four representative systems: UCSD Causal Copilot~\cite{wang2025causal}, MATMCD~\cite{shen2024exploring}, IBM AssetOpsBench~\cite{patel2025assetopsbench}, and SmartPilot~\cite{shyalika2025smartpilot_1} (Table~\ref{tab:summary_comparison}). UCSD and MATMCD primarily address offline causal graph refinement or simulated fault analysis with single or small sets of agents and fixed pipelines, while AssetOpsBench focuses on multi-agent simulation for asset-management scenarios rather than operator-facing decision support. SmartPilot offers a neurosymbolic multi-agent copilot, but its interaction patterns are mostly predefined and not grounded in standardized agentic protocols. CausalPulse is designed as an industry-grade multi-agent copilot that (i) implements a full end-to-end causal diagnostics pipeline (from preprocessing and anomaly detection through causal discovery, RCA, and recommendation), (ii) leverages standardized protocols for dynamic agent discovery, context-aware replanning, and real-time inter-agent communication, and (iii) is currently undergoing deployment on a live manufacturing line with an operator-facing dashboard.

\section{Pathway to Deployment}
\label{sec:deployment}

\begin{figure}[!ht]
\centering
\vspace{-4 mm}
\includegraphics[width=0.999\linewidth]{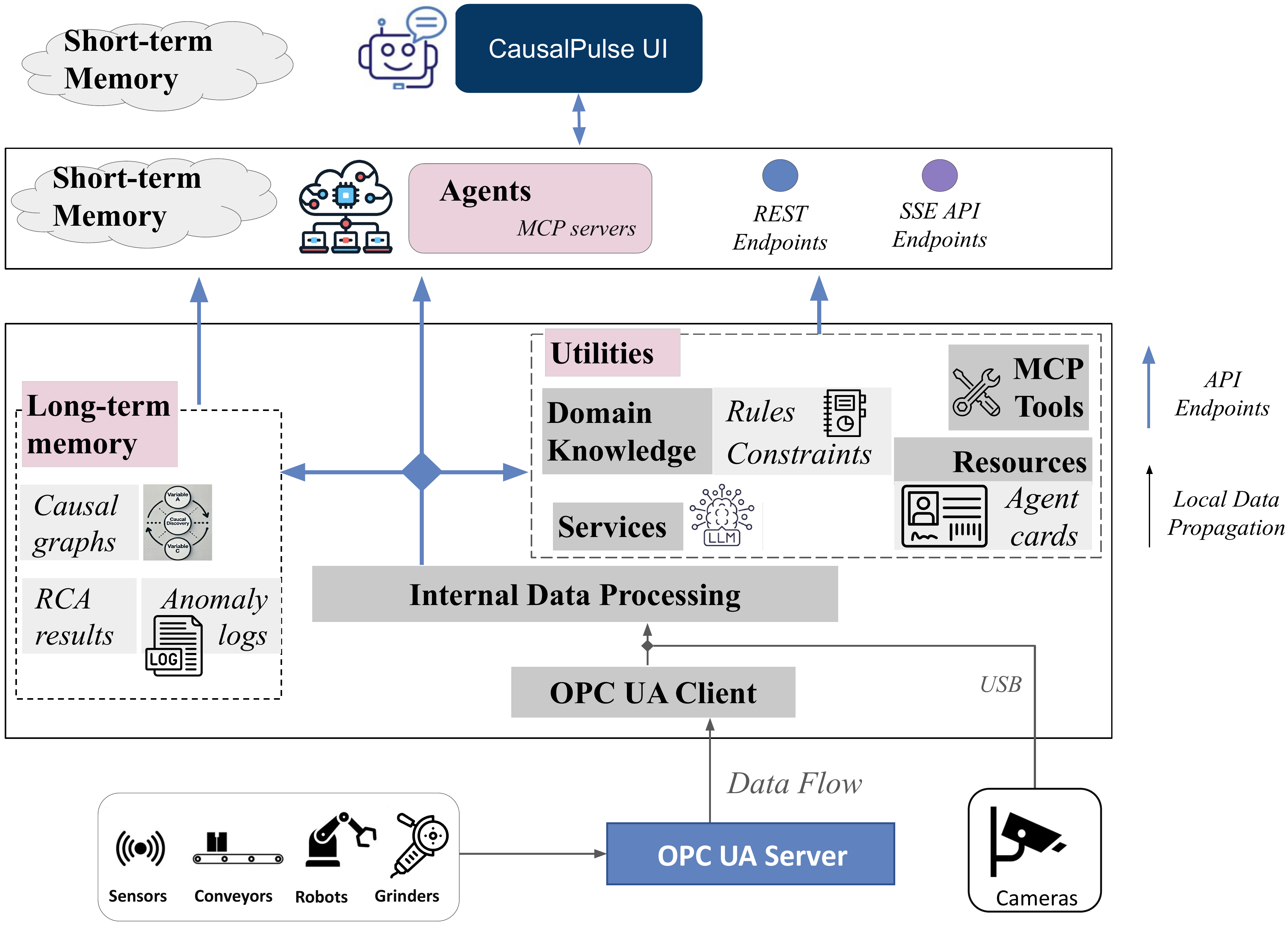}
  \vspace{-6 mm}
\caption{End-to-end deployment pipeline of CausalPulse}
  \label{fig:deployment}
\end{figure}

CausalPulse is currently being deployed on a Bosch production line, with the full deployment pipeline shown in Figure~\ref{fig:deployment}. At the factory level, an \emph{OPC UA Server (Open Platform Communications Unified Architecture)} aggregates live data from heterogeneous sources. The \emph{OPC UA Client} of CausalPulse receives and processes these streams, standardizing and structuring signals for agent-level consumption. The data are then integrated into the \emph{utility layer}, which hosts and retrieves domain knowledge, rule-based constraints, MCP tools, agent cards, and long-term memory. These are persistently accessed through standardized APIs and invoked by agents for real-time analytics and coordination. The \emph{agent layer} hosts agents as MCP servers and exposes REST and SSE API endpoints, enabling both human operators and other digital agents to interact with services. The short-term memory resides between the UI and agent layers, caching transient session-level data such as current workflow states, variable contexts, and intermediate results to enable fast context sharing during live interactions.  \emph{CausalPulse UI} is a browser-based interface that connects to the backend through FastAPI and LangGraph orchestration. The architecture supports local data propagation with minimal latency, while maintaining traceability and transparency across all layers. 

From the ongoing deployment experience, we identify key challenges in integrating agentic reasoning into safety-critical manufacturing pipelines. These include ensuring protocol interoperability across heterogeneous systems, calibrating operator trust through explainable outputs, and maintaining latency control under real-time production constraints. These challenges have directly informed the design of CausalPulse’s modular architecture and protocol-first orchestration strategy, which together enable safe and reliable industrial deployment. Ongoing work focuses on extending these capabilities to distributed, edge-enabled configurations for even greater scalability and responsiveness.

\section{Conclusion and Future Work}

CausalPulse is an industry-grade neurosymbolic multi-agent copilot for causal diagnostics, with ongoing extensions to include an Optimization Agent that enables automated, explainable process interventions. Future work focuses on edge deployment for low-latency distributed operations and reinforcement learning–based adaptation for autonomous calibration of thresholds and causal priors.

\bibliography{aaai2026}

@article{dawoud2025prorca,
  title={ProRCA: A Causal Python Package for Actionable Root Cause Analysis in Real-world Business Scenarios},
  author={Dawoud, Ahmed and Talupula, Shravan},
  journal={arXiv preprint arXiv:2503.01475},
  year={2025}
}

@book{spirtes2000causation,
  title={Causation, prediction, and search},
  author={Spirtes, Peter and Glymour, Clark N and Scheines, Richard},
  year={2000},
  publisher={MIT press}
}

@article{chickering2002optimal,
  title={Optimal structure identification with greedy search},
  author={Chickering, David Maxwell},
  journal={Journal of machine learning research},
  volume={3},
  number={Nov},
  pages={507--554},
  year={2002}
}

@inproceedings{shyalika2025nsfmap,
  title     = {NSF-MAP: Neurosymbolic Multimodal Fusion for Robust and Interpretable Anomaly Prediction in Assembly Pipelines},
  author    = {Shyalika, Chathurangi and Prasad, Renjith and El Kalach, Fadi and Venkataramanan, Revathy and Zand, Ramtin and Harik, Ramy and Sheth, Amit},
  booktitle = {Proceedings of the 34\textsuperscript{th} International Joint Conference on Artificial Intelligence (IJCAI) 2025},
  year      = {2025},
  note      = {Special Track on AI4Tech: AI Enabling Critical Technologies; currently available as an arXiv preprint arXtt{arXiv:2505.06333}},
  archivePrefix = {arXiv},
  eprint    = {2505.06333},
  primaryClass = {cs.LG},
}

@inproceedings{shyalika2025smartpilot_a,
  title={SmartPilot: A Multiagent CoPilot for Adaptive and Intelligent Manufacturing},
  author={Shyalika, Chathurangi and Prasad, Renjith and Al Ghazo, Alaa and Eswaramoorthi, Darssan and Kaur, Harleen and Muthuselvam, Sara Shree and Sheth, Amit},
  booktitle={2025 IEEE Conference on Artificial Intelligence (CAI)},
  pages={1--8},
  year={2025},
  organization={IEEE}
}

@inproceedings{shyalika2025smartpilot_b,
  title={SmartPilot: Agent-Based CoPilot for Intelligent Manufacturing},
  author={Shyalika, Chathurangi and Prasad, Renjith and Al Ghazo, Alaa and Eswaramoorthi, Darssan L and Shree Muthuselvam, Sara and Sheth, Amit},
  booktitle={Proceedings of the 24th International Conference on Autonomous Agents and Multiagent Systems},
  pages={3053--3055},
  year={2025}
}

@article{shen2024exploring,
  title={Exploring multi-modal integration with tool-augmented llm agents for precise causal discovery},
  author={Shen, ChengAo and Chen, Zhengzhang and Luo, Dongsheng and Xu, Dongkuan and Chen, Haifeng and Ni, Jingchao},
  journal={arXiv preprint arXiv:2412.13667},
  volume={1},
  number={3},
  year={2024}
}

@inproceedings{shyalika2025smartpilot_1,
  title={SmartPilot: Agent-Based CoPilot for Intelligent Manufacturing},
  author={Shyalika, Chathurangi and Prasad, Renjith and Al Ghazo, Alaa and Eswaramoorthi, Darssan L and Shree Muthuselvam, Sara and Sheth, Amit},
  booktitle={Proc. of the 24th International Conference on Autonomous Agents and Multiagent Systems},
  pages={3053--3055},
  year={2025}
}

@article{schick2023toolformer,
  title={Toolformer: Language models can teach themselves to use tools},
  author={Schick, Timo and Dwivedi-Yu, Jane and Dess{\`\i}, Roberto and Raileanu, Roberta and Lomeli, Maria and Hambro, Eric and Zettlemoyer, Luke and Cancedda, Nicola and Scialom, Thomas},
  journal={Advances in Neural Information Processing Systems},
  volume={36},
  pages={68539--68551},
  year={2023}
}

@article{kuo2021digital,
  title={Digital twin-enabled smart industrial systems: Recent developments and future perspectives},
  author={Kuo, Yong-Hong and Pilati, Francesco and Qu, Ting and Huang, George Q},
  journal={International Journal of Computer Integrated Manufacturing},
  volume={34},
  number={7-8},
  pages={685--689},
  year={2021},
  publisher={Taylor \& Francis}
}

@article{pietsch2024root,
  title={Root cause analysis in industrial manufacturing: A scoping review of current research, challenges and the promises of AI-driven approaches},
  author={Pietsch, Dominik and Matthes, Marvin and Wieland, Uwe and Ihlenfeldt, Steffen and Munkelt, Torsten},
  journal={Journal of Manufacturing and Materials Processing},
  volume={8},
  number={6},
  pages={277},
  year={2024},
  publisher={MDPI}
}

@inproceedings{yao2022react,
  title={React: Synergizing reasoning and acting in language models},
  author={Yao, Shunyu and Zhao, Jeffrey and Yu, Dian and Du, Nan and Shafran, Izhak and Narasimhan, Karthik R and Cao, Yuan},
  booktitle={The eleventh international conference on learning representations},
  year={2022}
}

@article{ko2023framework,
  title={A framework driven by physics-guided machine learning for process-structure-property causal analytics in additive manufacturing},
  author={Ko, Hyunwoong and Lu, Yan and Yang, Zhuo and Ndiaye, Ndeye Y and Witherell, Paul},
  journal={Journal of Manufacturing Systems},
  volume={67},
  pages={213--228},
  year={2023},
  publisher={Elsevier}
}

@article{pearl2019seven,
  title={The seven tools of causal inference, with reflections on machine learning},
  author={Pearl, Judea},
  journal={Communications of the ACM},
  volume={62},
  number={3},
  pages={54--60},
  year={2019},
  publisher={ACM New York, NY, USA}
}

@article{sofianidis2021review,
  title={A review of explainable artificial intelligence in manufacturing},
  author={Sofianidis, Georgios and Ro{\v{z}}anec, Jo{\v{z}}e M and Mladenic, Dunja and Kyriazis, Dimosthenis},
  journal={Trusted Artificial Intelligence in Manufacturing: A Review of the Emerging Wave of Ethical and Human Centric AI Technologies for Smart Production},
  pages={93--113},
  year={2021},
  publisher={Now}
}

@article{trivedi2024explainable,
  title={Explainable AI for industry 5.0: vision, architecture, and potential directions},
  author={Trivedi, Chandan and Bhattacharya, Pronaya and Prasad, Vivek Kumar and Patel, Viraj and Singh, Arunendra and Tanwar, Sudeep and Sharma, Ravi and Aluvala, Srinivas and Pau, Giovanni and Sharma, Gulshan},
  journal={IEEE Open Journal of Industry Applications},
  volume={5},
  pages={177--208},
  year={2024},
  publisher={IEEE}
}

@article{ribeiro2021describing,
  title={Describing structure and complex interactions in multi-agent-based industrial cyber-physical systems},
  author={Ribeiro, Luis and Gomes, Luis},
  journal={IEEE Access},
  volume={9},
  pages={153126--153141},
  year={2021},
  publisher={IEEE}
}

@article{harik2024analog,
  title={Analog and Multi-modal Manufacturing Datasets Acquired on the Future Factories Platform},
  author={Harik, Ramy and Kalach, Fadi El and Samaha, Jad and Clark, Devon and Sander, Drew and Samaha, Philip and Burns, Liam and Yousif, Ibrahim and Gadow, Victor and Tarekegne, Theodros and Saha, Nitol},
  journal={arXiv preprint arXiv:2401.15544},
  year={2024}
}

@misc{anthropic_mcp_2024,
  title        = {Introducing the Model Context Protocol},
  author       = {{Anthropic}},
  howpublished = {\url{https://www.anthropic.com/news/model-context-protocol}},
  month        = nov,
  year         = {2024},
  note         = {Accessed: 2025-10-15}
}

@misc{google_a2a_interoperability_2025,
  title        = {A2A: A New Era of Agent Interoperability},
  author       = {{Google A2A}},
  howpublished = {\url{https://developers.googleblog.com/en/a2a-a-new-era-of-agent-interoperability/}},
  year         = {2025},
note         = {Accessed: 2025-10-15}

}

@book{wooldridge2002multi,
  title={An introduction to multiagent systems},
  author={Wooldridge, Michael},
  year={2009},
  publisher={John wiley \& sons}
}

@inproceedings{bellifemine2000developing,
  title={Developing multi-agent systems with JADE},
  author={Bellifemine, Fabio and Poggi, Agostino and Rimassa, Giovanni},
  booktitle={International workshop on agent theories, architectures, and languages},
  pages={89--103},
  year={2000},
  organization={Springer}
}

@article{bandi2025rise,
  title={The Rise of Agentic AI: A Review of Definitions, Frameworks, Architectures, Applications, Evaluation Metrics, and Challenges},
  author={Bandi, Ajay and Kongari, Bhavani and Naguru, Roshini and Pasnoor, Sahitya and Vilipala, Sri Vidya},
  journal={Future Internet},
  volume={17},
  number={9},
  pages={404},
  year={2025},
  publisher={MDPI}
}

@article{bousetouane2025agentic,
  title={Agentic systems: A guide to transforming industries with vertical ai agents},
  author={Bousetouane, Fouad},
  journal={arXiv preprint arXiv:2501.00881},
  year={2025}
}

@article{gort2025agentedge,
  title={AgentEdge: Agentic AI for Service Orchestration in the Edge-Cloud Continuum},
  author={Gort, Berend and Kibalya, Godfrey M and Antonopoulos, Angelos},
  journal={Authorea Preprints},
  year={2025},
  publisher={Authorea}
}

@article{sawant2025unified,
  title        = {Unified, Modular, Self-Learning, Real-Time Agentic AI Ecosystem: Integrating Multi-Modal Data, Automation, Feedback Loops, and Secure Agent Orchestration},
  author       = {Sawant, Prashant D.},
  journal      = {Recent Trends in Artificial Intelligence \& Its Applications},
  volume       = {4},
  number       = {2},
  pages        = {53--61},
  year         = {2025},
  month        = {May--August},
  publisher    = {AI-Discovery Consultancy},
  issn         = {2583-4819},
  note         = {Founding Director, AI Research \& Director, AI-Discovery Consultancy, Melbourne, Australia}
}

@article{wang2025causal,
  title={Causal-copilot: An autonomous causal analysis agent},
  author={Wang, Xinyue and Zhou, Kun and Wu, Wenyi and Singh, Har Simrat and Nan, Fang and Jin, Songyao and Philip, Aryan and Patnaik, Saloni and Zhu, Hou and Singh, Shivam and others},
  journal={arXiv preprint arXiv:2504.13263},
  year={2025}
}

@article{patel2025assetopsbench,
  title={AssetOpsBench: Benchmarking AI Agents for Task Automation in Industrial Asset Operations and Maintenance},
  author={Patel, Dhaval and Lin, Shuxin and Rayfield, James and Zhou, Nianjun and Vaculin, Roman and Martinez, Natalia and O'donncha, Fearghal and Kalagnanam, Jayant},
  journal={arXiv preprint arXiv:2506.03828},
  year={2025}
}

@misc{agentcommprotocol2025,
  title        = {Agent Communication Protocol},
  howpublished = {\url{https://agentcommunicationprotocol.dev/introduction/welcome}},
  year         = {2025},
  note         = {Accessed: 2025-10-17}
}

@inproceedings{shyalika2025causaltrace,
  title={CausalTrace: A Neurosymbolic Causal Analysis Agent for Smart Manufacturing},
  author={Shyalika, Chathurangi and Sharma, Aryaman and Kalach, Fadi El and Jaimini, Utkarshani and Henson, Cory and Harik, Ramy and Sheth, Amit},
  booktitle={Proceedings of the AAAI Conference on Artificial Intelligence},
  volume={40},
  year={2026}
}

@article{derouiche2025agentic,
  title={Agentic AI Frameworks: Architectures, Protocols, and Design Challenges},
  author={Derouiche, Hana and Brahmi, Zaki and Mazeni, Haithem},
  journal={arXiv preprint arXiv:2508.10146},
  year={2025}
}

@inproceedings{wu2024autogen,
  title={Autogen: Enabling next-gen LLM applications via multi-agent conversations},
  author={Wu, Qingyun and Bansal, Gagan and Zhang, Jieyu and Wu, Yiran and Li, Beibin and Zhu, Erkang and Jiang, Li and Zhang, Xiaoyun and Zhang, Shaokun and Liu, Jiale and others},
  booktitle={First Conference on Language Modeling},
  year={2024}
}

@misc{crewai2025,
  title        = {CrewAI Documentation},
  author       = {CrewAI},
  year         = {2025},
  url          = {https://www.crewai.com/},
  note         = {Accessed: 2025-10-28}
}

@article{wang2024opendevin,
  title={Opendevin: An open platform for ai software developers as generalist agents},
  author={Wang, Xingyao and Li, Boxuan and Song, Yufan and Xu, Frank F and Tang, Xiangru and Zhuge, Mingchen and Pan, Jiayi and Song, Yueqi and Li, Bowen and Singh, Jaskirat and others},
  journal={arXiv preprint arXiv:2407.16741},
  volume={3},
  year={2024}
}

@misc{langgraph_2024,
  title        = {LangGraph: A Framework for Building Agentic Applications},
  author       = {{LangChain AI}},
  year         = {2024},
  howpublished = {\url{https://www.langchain.com/langgraph}},
  note         = {Accessed: 2025-10-17},
}

@article{marro2024scalable,
  title={A scalable communication protocol for networks of large language models},
  author={Marro, Samuele and La Malfa, Emanuele and Wright, Jesse and Li, Guohao and Shadbolt, Nigel and Wooldridge, Michael and Torr, Philip},
  journal={arXiv preprint arXiv:2410.11905},
  year={2024}
}

@misc{causallearn_2024,
  title        = {CausalLearn Documentation},
  author       = {{CausalLearn}},
  year         = {2024},
  howpublished = {\url{https://causal-learn.readthedocs.io/en/latest/}},
  note         = {Accessed: 2025-10-17}
}

@misc{pgmpy_2024,
  title        = {pgmpy: Probabilistic Graphical Models using Python},
  author       = {{pgmpy}},
  year         = {2024},
  howpublished = {\url{https://pgmpy.org/index.html}},
  note         = {Accessed: 2025-10-17}
}


\end{document}